\documentclass[sigconf,nonacm]{acmart}

\setcopyright{none}
\begin{document}

\title{Faster but Different: Diagnosing and Controlling Content Drift in Accelerated Multimodal Diffusion Language Models}

\author{Yaoxuan Dou}
\affiliation{%
  \institution{School of Mathematics and Statistics, Beijing Institute of Technology}
  \city{Beijing}
  \country{China}
}
\email{1120231803@bit.edu.cn}

\author{Yang Shu}
\authornote{Corresponding author.}
\affiliation{%
  \institution{Zhejiang University}
  \city{Hangzhou}
  \country{China}
}
\email{shuyang@zju.edu.cn}

\hypersetup{pdfauthor={Yaoxuan Dou and Yang Shu}}

\begin{abstract}
Training-free acceleration makes diffusion-based multimodal large language models (dMLLMs) more deployable, but it may silently change generated content. We study this serving-time consistency problem on 300 real images, comparing Fast-dLLM outputs with the same model's unaccelerated outputs. Across the mild parallelism induced in our long-form setting (1.05--1.25 committed tokens per step), confidence-threshold tuning changes decoding behavior but not baseline agreement. State-refresh ablations and an image-swap intervention instead identify stale visual and generated-text states as contributors to drift. For the tested Fast-dLLM implementation, shortening the KV-cache refresh interval yields a monotonic speed--agreement frontier and near-exact agreement at a measured $1.3\times$ speedup. The initial diagnosis also appears with dLLM-Cache and LaViDa, although dLLM-Cache recovers agreement only after both caches are tightened, which removes its speed advantage. Independent prompts and images reproduce the threshold-insensitivity and refresh recovery. A targeted audit finds genuine content substitution in half of 50 low-agreement pairs. In a separate blinded two-annotator evaluation, the pooled accelerated-minus-baseline factual-error difference is $0.00$ (95\% CI $[-0.17,+0.17]$); this sample detects no difference but does not establish factual equivalence. Finally, none of the tested adaptive or smoothed-refresh variants beats the fixed interval at matched compute. Our contribution is a paired diagnostic and an implementation-scoped consistency control, not an accuracy or safety guarantee.
\end{abstract}



\maketitle

\section{Introduction}

Multimodal assistants increasingly sit between users and visually rich information: screenshots, product pages, advertisements, news images, and social-media content. Realistic Web-agent benchmarks already require models to combine visual page understanding with downstream actions~\cite{visualwebarena2024,seeact2024}. In these latency-sensitive settings, a generated description or structured extraction may become evidence for a downstream search, content-analysis, or agentic decision. Acceleration is therefore not only a throughput concern. If the serving configuration silently changes what the model claims is visible, it can make the same content yield inconsistent intermediate evidence even when aggregate task accuracy appears stable. We study this inference-time information-integrity problem through diffusion-based multimodal large language models (dMLLMs), a new non-autoregressive family whose block-parallel generation makes aggressive acceleration especially attractive.

Unlike conventional left-to-right generation, dLLMs iteratively denoise a fully masked sequence through confidence-guided unmasking~\cite{nie2025llada}; LLaDA-V~\cite{you2025lladav} extends this paradigm with a vision encoder. The practical appeal is undercut by slow inference and dependency violations under parallel decoding~\cite{parallelbench2025,apd2025}. Fast-dLLM~\cite{wu2025fastdllm} addresses both through approximate block-wise KV caching and confidence-aware parallel decoding. Reported speedups reach an order of magnitude or more, and the confidence threshold is presented -- and used in follow-up work such as VRCD~\cite{vrcd2026} -- as the natural dial for trading inference speed against output quality.

\begin{figure}[t]
  \centering
  \includegraphics[width=\linewidth]{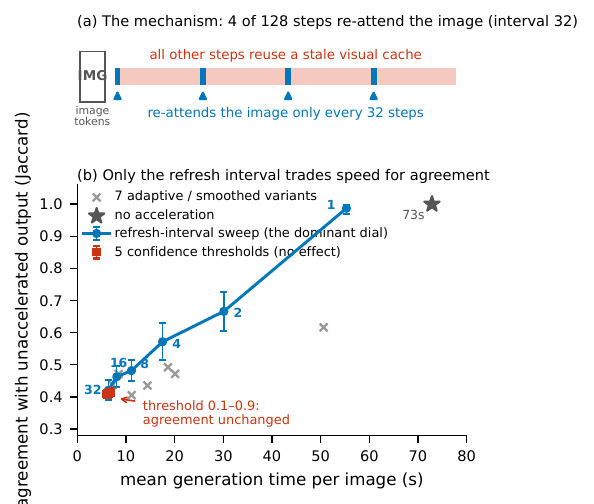}
  \caption{Why acceleration changes what the model says, and the dial that controls it in the settings we study. (a) Fast-dLLM's approximate KV cache re-attends to the image only once every \texttt{prefix\_refresh\_interval} steps (default 32): 124 of 128 denoising steps use cached visual states. (b) The threshold sweep uses 300 MME images; refresh and variant sweeps use a paired 50-image subset (means with 95\% bootstrap CIs). Threshold (red) leaves agreement unchanged; shrinking the refresh interval (blue, $32 \to 1$) trades speed for agreement continuously; no adaptive or smoothed-refresh variant (grey $\times$) beats this frontier. Agreement is against the unaccelerated output, not ground truth.}
  \Description{Panel a contrasts full refresh steps, which attend to image and text tokens, with incremental steps that reuse cached states. Panel b plots generation time against agreement: threshold settings form a low-agreement cluster, refresh settings trace an increasing frontier, and adaptive variants lie below it.}
  \label{fig:teaser}
\end{figure}

This threshold-as-dial assumption has, to our knowledge, never been directly tested for multimodal, open-ended generation, where the failure mode of interest is not a drop in task accuracy but a change in \emph{what the model claims is in the image}. We test it directly. Using LLaDA-V and Fast-dLLM on 300 real images spanning diverse content categories, we generate detailed image descriptions under six configurations -- an unaccelerated baseline and five confidence thresholds spanning the method's operating range -- and measure \emph{baseline agreement}: how closely the accelerated output matches, in content, the unaccelerated output for the same image. Two clarifications frame everything that follows. First, baseline agreement is a \emph{consistency} construct, not an accuracy construct: it quantifies whether acceleration changes what the model says, not whether either version is correct about the image; Section~\ref{sec:audit} separately evaluates image-grounded factuality and detects no accelerated--baseline difference in a limited 50-image study, without establishing equivalence. Second, in the released implementation the threshold's effective direction is the reverse of its intuitive reading -- larger values commit \emph{more} tokens per step (Section~\ref{sec:rq1}). We find that the threshold-as-dial assumption does not hold in this tested range: once acceleration is enabled, content diverges from the unaccelerated output by a large, nearly constant amount regardless of threshold. We call this an \emph{off-switch effect}: turning acceleration on moves the model onto a different generation trajectory whose distance from the unaccelerated trajectory is not controlled by the tested thresholds.

We do not stop at the diagnosis. Inspecting Fast-dLLM, we trace the effect to its KV-cache schedule: full recomputation occurs only at a fixed refresh interval, with intermediate steps reusing cached visual and generated-text states (Figure~\ref{fig:teaser}a). Orthogonal refresh ablations show that both state groups contribute, with visual-state refresh recovering more agreement; a direct image-swap intervention shows that incremental steps cannot incorporate a changed image until a full refresh. This is a scheduling-induced analogue of the gradual visual-information loss reported for autoregressive VLMs~\cite{visualsteering2025}. Guided by this mechanism, we show that the refresh interval -- not the threshold -- is the dominant observed consistency control for the tested Fast-dLLM implementation: sweeping it produces a continuous, monotonic agreement curve, and its most conservative setting nearly eliminates drift while retaining a real speedup.

To assess how much of this story is specific to one acceleration implementation, we repeat the full refresh-interval sweep using an independent, differently-engineered caching method, dLLM-Cache~\cite{liu2025dllmcache}. The initial diagnosis recurs: the two implementations produce similar drift under their default configurations. The single-parameter remedy does not transfer: dLLM-Cache's drift stays nearly flat when only its prompt cache is refreshed more often, pointing to its independently scheduled generated-token cache as a second candidate source. A 15-image joint pilot supports this interpretation, but recovering agreement by tightening both caches removes dLLM-Cache's speed advantage. We report this as an implementation boundary condition. Finally, because our core evidence relies on lexical similarity, we manually audit 50 low-similarity generations and find that half reflect genuine content substitution, including invented content, rather than harmless paraphrase.

For Web-facing multimodal systems, the central contribution is a serving-time diagnostic that detects when an efficiency mechanism changes the information extracted from visual content, separately from whether either output is factually correct. Concretely, we contribute: (1) systematic evidence that confidence-threshold tuning does not control content drift across the induced parallelism range, using 300 images, two additional output formats on an independent sample, and step-level instrumentation; (2) causal evidence, in the tested Fast-dLLM implementation, that KV-cache staleness contributes to drift, from orthogonal state-refresh ablations and an image-swap intervention; (3) a continuously tunable refresh control with a measured efficiency--agreement frontier; (4) cross-implementation, cross-model, and matched-latency checks that scope the diagnosis and remedy; (5) manual and blinded studies separating trajectory consistency from image-grounded factuality; and (6) a negative-results study showing that no tested adaptive or smoothed-refresh variant beats the fixed interval at matched compute.

\section{Related Work}

\begin{sloppypar}
\textbf{Accelerating diffusion language model inference.} Fast-dLLM~\cite{wu2025fastdllm} is the primary object of our study, combining block-wise approximate KV caching with confidence-aware parallel decoding to accelerate LLaDA-family models by an order of magnitude or more. Fast-dLLM++~\cite{kasa2026fastdllmpp} subsequently replaces weakest-token selection with confidence-profile decoding while leaving the cache implementation unchanged; it improves the text-only accuracy--throughput frontier but does not study multimodal cache-refresh consistency. dLLM-Cache~\cite{liu2025dllmcache} pursues a related but independently engineered goal, combining long-interval prompt caching with partial, similarity-guided updates to the response cache. Adaptive Parallel Decoding (APD)~\cite{apd2025} instead uses a small auxiliary autoregressive model to decide how many tokens to commit. These works evaluate task accuracy or throughput rather than sensitivity of open-ended multimodal content to exposed serving parameters.
\end{sloppypar}

\begin{sloppypar}
\textbf{Quality degradation under parallel decoding.} ParallelBench~\cite{parallelbench2025} is closest in spirit to our diagnosis: it introduces a benchmark to quantify quality degradation under parallel decoding in dLLMs, and attributes the degradation to the conditional-independence assumption violating token dependencies. However, ParallelBench evaluates purely text-only dLLMs and does not study the multimodal setting, nor does it examine cache-refresh scheduling as a distinct variable from decoding parallelism. Our work extends this line of inquiry to multimodal generation and isolates a mechanism (cache staleness) that is orthogonal to the dependency-violation account.
\end{sloppypar}

\begin{sloppypar}
\textbf{Visual token handling in diffusion MLLMs.} A concurrent line of work studies how diffusion MLLMs allocate attention across visual tokens during parallel decoding. VRCD~\cite{vrcd2026} proposes redundancy-aware reranking of co-selected high-confidence tokens within a single decoding step; a broader study of visual token redundancy~\cite{visualredundancy2026} targets efficiency-accuracy trade-offs on standard VQA benchmarks. Neither analyzes cache-refresh scheduling, nor content divergence from a paired unaccelerated reference on open-ended multimodal generation -- the specific analysis we present.
\end{sloppypar}

\textbf{Hallucination and visual grounding in VLMs.} Visual Information Steering~\cite{visualsteering2025} reports gradual loss of visually grounded token preference during autoregressive generation. Our image-swap intervention identifies a different, deterministic mechanism: Fast-dLLM's incremental steps reuse cached visual states and cannot incorporate a changed image until refresh. We therefore use the autoregressive finding only as a structural analogy, not as evidence that the two mechanisms are identical.

\textbf{Multimodal agents on the Web.} VisualWebArena~\cite{visualwebarena2024} evaluates agents on realistic visually grounded Web tasks, while SeeAct~\cite{seeact2024} combines visual understanding and action grounding on websites. These works motivate why stable visual evidence can matter downstream. Our experiments do not evaluate Web navigation or task success; they isolate a lower-level serving consistency question that such systems may encounter.

\textbf{Positioning summary.} Table~\ref{tab:related} summarizes how our work differs from the six most closely related papers. VRCD and ParallelBench are recent arXiv preprints, so we treat them as concurrent work and read novelty claims with that caveat. To the best of our knowledge, no prior work combines a multimodal setting with a controlled sweep of both confidence threshold and cache refresh interval against a paired unaccelerated reference, or tests the proposed control across two independently engineered caching implementations.

\begin{table}[t]
  \caption{How this paper differs from the closest related work.}
  \label{tab:related}
  \small
  \setlength{\tabcolsep}{4pt}
  \begin{tabular}{lccccc}
    \toprule
     & MM & Thresh. & Refresh & Cross-impl. & Content \\
     &    & vs.\ drift & vs.\ drift & check & metric \\
    \midrule
    Fast-dLLM~\cite{wu2025fastdllm}      & $\times$ & $\times$ & $\times$ & $\times$ & $\times$ \\
    dLLM-Cache~\cite{liu2025dllmcache}   & $\times$ & $\times$ & $\times$ & $\times$ & $\times$ \\
    APD~\cite{apd2025}                    & $\times$ & $\times$ & $\times$ & $\times$ & $\times$ \\
    ParallelBench~\cite{parallelbench2025}& $\times$ & \checkmark & $\times$ & $\times$ & \checkmark \\
    VRCD~\cite{vrcd2026}                  & \checkmark & $\times$ & $\times$ & $\times$ & $\times$ \\
    Vis.\ redund.~\cite{visualredundancy2026}& \checkmark & $\times$ & $\times$ & $\times$ & $\times$ \\
    \textbf{Ours}                         & \checkmark & \checkmark & \checkmark & \checkmark & \checkmark \\
    \bottomrule
  \end{tabular}
\end{table}

\section{Background}

\subsection{Diffusion Language Model Generation}
Unlike autoregressive models, LLaDA-family models~\cite{nie2025llada,you2025lladav} generate a sequence of \texttt{gen\_length} tokens by initializing them all to a mask token and iteratively replacing masked positions with predicted tokens over \texttt{steps} denoising steps, optionally divided into \texttt{block\_length}-sized blocks for semi-autoregressive block-wise generation. At each step, the model computes a full forward pass, and a remasking policy (e.g., low-confidence remasking) decides which currently-masked positions to commit based on the model's predicted confidence.

\subsection{Fast-dLLM Acceleration}
Fast-dLLM~\cite{wu2025fastdllm} accelerates this process with two mechanisms. First, an approximate block-wise KV cache avoids recomputing attention for the prompt and already-completed blocks at every step: a full forward pass (recomputing keys and values for the entire sequence, including image tokens) is performed only when the current step index is a multiple of \texttt{prefix\_refresh\_interval}; all other steps perform an incremental forward pass restricted to the active generation block, reusing cached keys and values for everything before it. Second, confidence-aware parallel decoding replaces the fixed per-step token budget with a threshold rule: any masked position whose predicted confidence exceeds \texttt{threshold} is committed in the current step, allowing a variable, potentially large number of tokens to be unmasked per step when the model is confident.

\subsection{dLLM-Cache}
\begin{sloppypar}
dLLM-Cache~\cite{liu2025dllmcache} is an independently engineered caching method for the same class of models. It separately controls the refresh interval for the prompt cache (\texttt{prompt\_interval\_steps}) and for the response (already-generated-token) cache (\texttt{gen\_interval\_steps}), combined with a similarity-based partial update rule (\texttt{transfer\_ratio}) for deciding which cached features to refresh between full recomputes. We use it in Section~\ref{sec:generalization} as an independent implementation to test the generality of our diagnosis.
\end{sloppypar}

\section{Experimental Setup}

\textbf{Model and data.} We use LLaDA-V (\texttt{GSAI-ML/LLaDA-V}), an 8B-parameter dMLLM built on a LLaDA-8B language backbone with a SigLIP2-SO400M vision encoder. We draw 300 images with a fixed random seed from the MME benchmark~\cite{fu2023mme}; RQ1 uses all 300, while the more computationally intensive mechanism, refresh, and cross-implementation studies use the same fixed 50-image subset. The primary prompt (``Please describe the image in detail.'') uses \texttt{steps=gen\_length=block\_length=128}. A prompt/sample-generalization check draws 50 nonoverlapping images with seed 20260727 and uses a one-sentence prompt (length/steps 32) and fixed-schema JSON prompt (length/steps 96). An auxiliary cross-model check uses the official LaViDa checkpoint and code~\cite{lilavida} on the primary 50 rows and prompt, sweeping its exposed prefix-cache switch and denoising NFE ratio.

\begin{sloppypar}
\textbf{Metrics and terminology.} For each accelerated configuration, we compare the generated text against the unaccelerated (\texttt{use\_fast\_dllm=False}) generation for the same image, computed once per image and reused as a common reference across all configurations. We call the resulting quantity \emph{baseline agreement} and its complement \emph{content drift}. Two distinct constructs must not be conflated here, and we keep them separate throughout. \emph{Baseline agreement} (this section's metrics) measures whether acceleration changes what the model says relative to its own unaccelerated behavior -- a consistency and reproducibility property. \emph{Image-grounded factuality} measures whether a description is correct about the image -- an accuracy property that requires looking at the image, which no text-to-text metric can assess, and which we probe separately via a manual audit and a ground-truth spot-check in Section~\ref{sec:audit}. High agreement does not imply factual correctness (the unaccelerated output can itself hallucinate), and low agreement does not per se imply error (it may be benign paraphrase). We report (i) wall-clock generation time; (ii) word-level Jaccard similarity between the accelerated and unaccelerated text (computed over lowercased alphabetic word sets); (iii) distinct-2 ratio; and (iv) repeat-3 rate. Unless stated otherwise, 95\% CIs are paired percentile-bootstrap intervals over images (10,000 resamples, seed 0). Section~\ref{sec:audit} corroborates Jaccard with BERTScore and manual evaluation.
\end{sloppypar}

\textbf{Reproducibility.} Primary experiments use one NVIDIA RTX 6000 Ada GPU (48GB), PyTorch 2.1.2/CUDA 12.1, and the released LLaDA-V code with Fast-dLLM and dLLM-Cache hooks. Images are selected with recorded seeds; comparisons are paired by image. Decoding is deterministic (no temperature or sampling), so drift is attributable to configuration, not run-to-run randomness. Times exclude loading and preprocessing. The LaViDa check uses its official repository (commit \texttt{24220c0}) and checkpoint in a separate environment on the same GPU. We additionally repeat eight timing configurations 30 times each on an A800-SXM4-80GB, with warm-up, synchronized timing, and globally randomized order: the no-acceleration/default times are 71.5/6.3s ($11.3\times$), matching 71.9/6.4s ($11.3\times$) on the Ada GPU, and every qualitative speed claim transfers. Table~\ref{tab:ledger} itemizes all 4{,}930 generations. We will release scripts, raw logs, image identifiers, and analysis code upon publication.

\section{RQ1: Is Content Drift Threshold-Sensitive?}
\label{sec:rq1}

\begin{sloppypar}
We generate all 300 images under six configurations: an unaccelerated baseline, and Fast-dLLM with \texttt{threshold} $\in \{0.9, 0.7, 0.5, 0.3, 0.1\}$. In the released implementation, a masked position is committed early when confidence exceeds $1 - \texttt{threshold}/(n{+}1)$; larger values therefore lower the bar and commit more tokens. Instrumentation confirms that \texttt{threshold=0.9} is most aggressive ($1.254$ tokens/step) and \texttt{threshold=0.1} most conservative ($1.055$) -- the reverse of an intuitive reading. This sweep yields 1,800 real generations.
\end{sloppypar}

\begin{figure}[t]
  \centering
  \includegraphics[width=\linewidth]{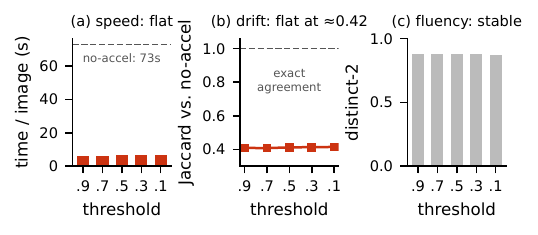}
  \caption{Threshold sweep (300 images $\times$ 6 configurations, 1,800 generations; 95\% paired-bootstrap CIs). (a) Acceleration yields a 10--12$\times$ speedup. (b) Agreement remains flat at $\approx$0.41--$0.42$. (c) distinct-2 remains stable.}
  \Description{Three plots compare the unaccelerated baseline with five threshold settings. Generation time drops sharply for every accelerated setting, while Jaccard agreement and distinct-2 remain nearly flat across thresholds.}
  \label{fig:threshold}
\end{figure}

Figure~\ref{fig:threshold} summarizes the result. Acceleration reduces mean generation time from 72.9s to 6.1--6.9s (a 10--12$\times$ speedup). Mean Jaccard lies in the narrow range $0.411$--$0.417$. A paired Friedman test finds no threshold effect ($\chi^2(4)=1.36$, $p=0.85$); all 10 pairwise contrasts remain nonsignificant after Holm correction. The extreme comparison (threshold $0.1-0.9$) is only $+0.0056$ (95\% paired bootstrap CI $[-0.0025,+0.0139]$, Wilcoxon $p=0.75$). Using a review-stage equivalence region of $\pm0.02$ Jaccard (declared after, not before, the original experiment), its 90\% CI $[-0.0012,+0.0125]$ lies wholly inside the region and paired TOST rejects nonequivalence ($p=0.00034$); paired $d_z=0.078$ and Cliff's $\delta=0.017$ are negligible. Thus the data bound the effect rather than merely failing to reject a large one. distinct-2 remains $0.873$--$0.877$ and repeat-3 at $0$--$0.3\%$: outputs are fluent but describe different content.

\textbf{Is the threshold actually operative?} A skeptical reading of this flatness is that the threshold simply was not doing anything in our pipeline -- a plumbing failure rather than a finding. We rule this out by re-running the full sweep (250 generations) through an instrumented decoding path that logs per-step commit counts and confidences while leaving the computation untouched. Three facts result. First, the instrumented runs reproduce the original outputs byte-for-byte in 250/250 cases, so the instrumentation observes the real code path with zero perturbation. Second, the threshold demonstrably and monotonically changes decoding behavior: mean steps per generation fall from $121.4$ (threshold $0.1$) to $102.8$ (threshold $0.9$), tokens per step rise from $1.055$ to $1.254$, and the fraction of multi-token steps rises from $4.5\%$ to $17.3\%$ (Wilcoxon on paired step counts, $p<10^{-4}$); outputs at different thresholds differ on ${\sim}95\%$ of image pairs. Third, despite this verified behavioral change, agreement stays flat at $0.41$--$0.42$. The dial turns; agreement does not move. We accordingly scope the claim: the threshold induces only a mild parallelism range here ($1.05$--$1.25$ tokens/step), and our insensitivity result is a statement about that range -- which is precisely the method's operating range on this workload at default settings. A representative case: a painting the baseline describes as ``a Rembrandt-style street scene ... a man on a horse'' becomes, in every accelerated configuration, ``an ancient Roman courtyard ... ruins of a grand building'' -- a coherent but different scene, not a corrupted one.

We term this the \emph{off-switch effect}: enabling Fast-dLLM acceleration is better modeled as switching to a qualitatively different generation trajectory than as smoothly interpolating along a single quality-speed curve controlled by threshold. This directly contradicts the implicit deployment assumption -- shared by Fast-dLLM's own presentation and by follow-up work such as VRCD~\cite{vrcd2026} -- that the confidence threshold is the knob controlling how much acceleration changes the output. The claim holds at default and practical cache-refresh settings, within the parallelism range induced here; Section~\ref{sec:interaction} maps the one boundary regime (near per-step refresh) where a small threshold interaction emerges.

\textbf{Prompt and sample generalization.} On 50 new MME images (zero overlap with the primary 300), sentence/JSON prompts give default-cache agreement 0.592/0.492. Threshold extremes remain indistinguishable (differences $-0.0077/+0.0064$; $p=0.18/0.86$); the sentence contrast meets the review-stage $\pm0.02$ TOST criterion, while JSON does not establish equivalence. Per-step refresh recovers agreement to 0.974/0.921 at $1.31\times/1.82\times$ speedups; sentence/valid-JSON compliance remains 96--98\%/90--96\%. Thus both prompts replicate the diagnosis and remedy, with the structured result limited to no detected threshold effect.

\subsection{Does Drift Vary by Visual Content Type?}
\label{sec:category}

MME organizes its images into content categories. Our 50-image sample spans 11 of them, unevenly (1--10 images per category, Table~\ref{tab:category}), so we treat this breakdown as exploratory; rows with $\leq 2$ images are included for completeness only. A pattern nevertheless emerges: \emph{artwork} has the lowest average agreement (0.317), consistent with the Roman-street-painting example, where the model can produce different but plausible compositional details. \emph{Scene} and \emph{posters}, with a few salient objects, score higher (0.450--0.453). This direction is consistent with, but does not by itself establish, our mechanistic account.

\begin{table}[t]
  \caption{Content drift by MME category (exploratory). ``Images'' is the number of distinct images per category; ``Gen.'' is the number of accelerated generations ($=5\times$ Images). Rows with Images $\leq 2$ are not statistically reliable.}
  \label{tab:category}
  \small
  \begin{tabular}{lccc}
    \toprule
    Category & Images & Gen. & Avg. Jaccard \\
    \midrule
    artwork               & 7  & 35 & 0.317 \\
    OCR                   & 1  & 5  & 0.337 \\
    code\_reasoning       & 1  & 5  & 0.358 \\
    celebrity             & 6  & 30 & 0.390 \\
    commonsense\_reasoning& 4  & 20 & 0.403 \\
    existence             & 2  & 10 & 0.410 \\
    landmark              & 8  & 40 & 0.422 \\
    count                 & 2  & 10 & 0.442 \\
    posters               & 8  & 40 & 0.450 \\
    scene                 & 10 & 50 & 0.453 \\
    position               & 1  & 5  & 0.603 \\
    \bottomrule
  \end{tabular}
\end{table}

\section{RQ2: What Causes the Drift, and Can It Be Fixed?}

\subsection{Mechanism}
Inspecting the Fast-dLLM implementation, we find that its approximate KV cache is refreshed on a fixed schedule: a full forward pass over the entire sequence -- including image tokens -- occurs only when the current denoising step index is a multiple of \texttt{prefix\_refresh\_interval} (default 32); all other steps perform an incremental forward pass restricted to the generation block, reusing cached key/value states for the prompt and image from the last refresh. With the default interval, only 4 of 128 steps involve genuinely re-attending to the image; the remaining 124 steps generate conditioned on an increasingly stale visual representation.

This is structurally analogous to gradual visual-information loss reported for autoregressive VLMs~\cite{visualsteering2025}. Here, however, the effect is a deterministic cache window in which incremental steps reuse image states and therefore cannot incorporate newly supplied visual information until the next full refresh.

\textbf{Causal isolation.} At interval 4, refreshing only image-token states yields Jaccard $0.498$, while refreshing only generated-text states yields $0.454$ (paired difference $+0.044$, $p=0.028$); both improve over the default interval-32 reference ($0.424$), but neither reaches full interval-4 refresh ($0.572$). Visual staleness is therefore the larger contributor, not the sole one. In a separate intervention, we replace image A with image B after step 4. With no subsequent full refresh (interval 128), swapped and non-swapped outputs are byte-identical in 10/10 pairs; with per-step refresh, outputs instead move strongly toward B (Jaccard to A/B baselines $0.228/0.589$). This directly verifies that refresh scheduling gates access to changed visual evidence.

\subsection{The Refresh Interval Is the Dominant Dial}

\begin{figure}[t]
  \centering
  \includegraphics[width=\linewidth]{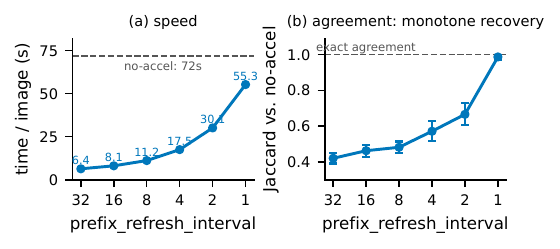}
  \caption{Speed and baseline agreement as a function of \texttt{prefix\_refresh\_interval} (50 images $\times$ 6 configurations, 300 generations; error bars are 95\% bootstrap CIs). Unlike threshold, the refresh interval produces a continuous, monotonic trade-off between speed and agreement with the unaccelerated output, converging to near-exact agreement at its most conservative setting while retaining a real speedup over the unaccelerated baseline (dashed line).}
  \Description{A two-axis plot shows that shorter refresh intervals increase both generation time and agreement. Agreement rises monotonically from about 0.42 at interval 32 to about 0.99 at interval 1, which remains faster than the unaccelerated baseline.}
  \label{fig:refresh}
\end{figure}

We fix \texttt{threshold=0.5} and sweep \texttt{prefix\_refresh\_interval} $\in \{32, 16, 8, 4, 2, 1\}$ on the same 50 images. Figure~\ref{fig:refresh} and Table~\ref{tab:refresh} show the result: unlike threshold, the refresh interval produces a clean, continuous, monotonic trade-off. Generation time increases from 6.4s to 55.3s as the interval shrinks from 32 to 1, while Jaccard similarity to the unaccelerated baseline increases monotonically from 0.420 to 0.987. This trend is highly significant (Page's $L$ test for the ordered alternative, $z=11.6$, $p<10^{-10}$). Four of five adjacent improvements survive Holm correction (adjusted $p=0.00031$, $0.0063$, $0.0135$, and $<10^{-7}$ for $32{\to}16$, $8{\to}4$, $4{\to}2$, and $2{\to}1$); only $16{\to}8$ is nonsignificant ($p_{\mathrm{Holm}}=0.247$). At \texttt{prefix\_refresh\_interval=1} -- full recomputation, including image attention, at every step -- 80\% of the 50 generations are word-for-word identical to the unaccelerated baseline. The configuration still runs $1.3\times$ faster than no acceleration at all (55.3s vs. 71.9s), because confidence-aware parallel decoding continues to commit multiple tokens per step whenever the model is confident, independent of cache freshness. Cache refresh frequency and confidence-driven parallelism are therefore two independent, separately controllable sources of speedup. Contrary to common practice, it is the former -- not the latter -- that determines agreement with the unaccelerated output in the regimes we study.

\begin{table}[t]
  \caption{Speed--agreement trade-off across refresh intervals (50 images each).}
  \label{tab:refresh}
  \small
  \begin{tabular}{lccc}
    \toprule
    \texttt{prefix\_refresh\_interval} & Time (s) & Jaccard & Exact match \\
    \midrule
    32 (default) & 6.37  & 0.420 & 0/50 \\
    16            & 8.12  & 0.463 & 0/50 \\
    8             & 11.16 & 0.482 & 0/50 \\
    4             & 17.53 & 0.572 & 2/50 \\
    2             & 30.11 & 0.666 & 3/50 \\
    1             & 55.29 & 0.987 & 40/50 \\
    \midrule
    No acceleration & 71.90 & 1.000 & 50/50 \\
    \bottomrule
  \end{tabular}
\end{table}

\subsection{Does Threshold Interact with Refresh Interval?}
\label{sec:interaction}

\begin{sloppypar}
Our threshold-insensitivity result (Section~\ref{sec:rq1}) was established at the default \texttt{prefix\_refresh\_interval=32}, and our refresh-interval result above was established at the default \texttt{threshold=0.5}; neither directly tests whether the two parameters interact away from these reference points, a concern a reviewer of an earlier draft of this work correctly raised. We test this with a $3\times3$ grid, \texttt{threshold} $\in \{0.9, 0.5, 0.1\} \times$ \texttt{prefix\_refresh\_interval} $\in \{32, 4, 1\}$, filling in the four off-diagonal cells not covered by our existing sweeps, on the same 50 images.
\end{sloppypar}

At the default refresh interval (32), threshold remains flat (mean Jaccard $0.419$, $0.420$, $0.406$ for threshold $0.9$, $0.5$, $0.1$), replicating Section~\ref{sec:rq1}. At refresh$=1$, a targeted exploratory contrast gives agreement $0.955$ at threshold$=0.9$ versus $0.995$ at threshold$=0.1$, a difference-in-differences of $-0.054$ relative to refresh$=32$ (paired Wilcoxon $p=0.0085$; 95\% bootstrap CI $[-0.092,-0.021]$); the analogous refresh$=4$ contrast is nonsignificant ($p=0.37$). However, a mixed-effects model on all 450 paired observations, $\text{agreement}\sim\text{threshold}\times\text{refresh}+(1\mid\text{image})$, finds no omnibus interaction (likelihood-ratio $\chi^2(4)=7.42$, $p=0.115$). We therefore do not claim a general interaction: the local near-ceiling contrast is hypothesis-generating, whereas the model-level result supports refresh interval as the dominant agreement lever.

\section{RQ3: Does This Generalize Across Implementations?}
\label{sec:generalization}

\begin{sloppypar}
A natural concern is that both the diagnosis and the remedy are artifacts of Fast-dLLM's specific implementation. We test this by repeating the refresh-interval sweep with dLLM-Cache~\cite{liu2025dllmcache}, an independently engineered caching method also integrated into the LLaDA-V codebase. dLLM-Cache exposes \texttt{prompt\_interval\_steps} as the direct analogue of \texttt{prefix\_refresh\_interval}, controlling how often the prompt/image cache is fully refreshed. We fix its other parameters at their released defaults (\texttt{gen\_interval\_steps=7} and \texttt{transfer\_ratio=0.25}) and sweep \texttt{prompt\_interval\_steps} over $\{25, 16, 8, 4, 2, 1\}$ on the same 50 images.
\end{sloppypar}

\begin{figure}[t]
  \centering
  \includegraphics[width=\linewidth]{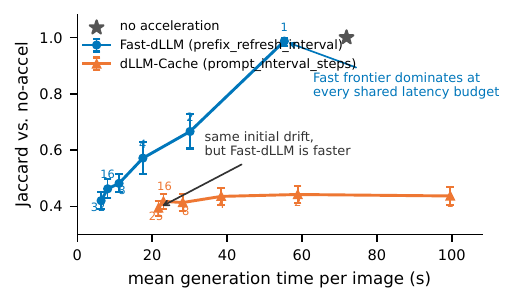}
  \caption{Measured latency--agreement frontiers for Fast-dLLM and dLLM-Cache (50 paired images per point; error bars are 95\% bootstrap CIs). At every shared latency budget, the best feasible Fast-dLLM setting has higher agreement; the unaccelerated baseline is shown separately.}
  \Description{Agreement is plotted against generation time for Fast-dLLM and dLLM-Cache. The Fast-dLLM curve rises toward the unaccelerated reference, whereas the dLLM-Cache prompt-refresh curve remains near 0.4 agreement as latency increases.}
  \label{fig:dllmcache}
\end{figure}

\begin{sloppypar}
Figure~\ref{fig:dllmcache} shows a split result. The \emph{diagnosis recurs}: dLLM-Cache's default configuration produces Jaccard 0.394, close to Fast-dLLM's 0.420, supporting a shared failure mode in these two tested approximate-caching implementations. The \emph{single-parameter remedy does not transfer}: as \texttt{prompt\_interval\_steps} shrinks from 25 to 1, Jaccard moves only from 0.394 to 0.437, with zero exact matches at any setting (Table~\ref{tab:dllmcache}). A paired Friedman test detects a small effect ($\chi^2(5)=22.4$, $p<0.001$), and the extreme comparison gives $+0.043$ (95\% CI $[0.015,0.070]$, Wilcoxon $p=0.001$), over an order of magnitude smaller than Fast-dLLM's $0.567$ range. More importantly, the actual-latency frontier removes any setting-index ambiguity: Fast-dLLM has higher agreement at all six shared dLLM-Cache latency budgets, by $0.153$--$0.550$. Even full per-step prompt-cache recomputation therefore falls far short of the unaccelerated reference.
\end{sloppypar}

\begin{sloppypar}
We interpret this as an open boundary condition. dLLM-Cache exposes a second, independently scheduled cache -- \texttt{gen\_interval\_steps}, refreshing the cache for \emph{already-generated} tokens -- held at its default (7) above; we hypothesized it is a distinct drift source that prompt-cache refresh alone cannot address. In a 15-image joint-sweep pilot (\texttt{prompt\_interval\_steps=1}; \texttt{gen\_interval\_steps} $\in \{7, 4, 2, 1\}$), Jaccard rises modestly from 7 to 2 ($0.450 \to 0.459 \to 0.506$), then reaches $0.945$ at 1, with 13/15 outputs word-for-word identical to the unaccelerated baseline. This pilot supports the generated-token cache as a second drift source, but its small size limits generalization. At both intervals equal to 1, average generation time is $109.0$s, slower than the $71.9$s unaccelerated baseline. Thus agreement is recoverable in the tested pilot, but only after eliminating the acceleration benefit.
\end{sloppypar}

\begin{table}[t]
  \caption{dLLM-Cache: initial drift replicates, remedy does not (50 images each).}
  \label{tab:dllmcache}
  \small
  \begin{tabular}{lccc}
    \toprule
    \texttt{prompt\_interval\_steps} & Time (s) & Jaccard & Exact match \\
    \midrule
    25 (default) & 21.66 & 0.394 & 0/50 \\
    16            & 23.01 & 0.418 & 0/50 \\
    8             & 28.17 & 0.414 & 0/50 \\
    4             & 38.44 & 0.436 & 0/50 \\
    2             & 58.86 & 0.442 & 0/50 \\
    1             & 99.58 & 0.437 & 0/50 \\
    \bottomrule
  \end{tabular}
\end{table}

\paragraph{Auxiliary cross-model check: LaViDa.}
\label{sec:lavida}

We run 300 paired generations on official LaViDa~\cite{lilavida}, crossing its prefix-cache switch with NFE. At matched NFE 128, caching cuts mean time from 21.33s to 4.48s ($4.77\times$) but lowers agreement with the uncached output to 0.299 (95\% CI [0.265, 0.334]); at NFE 64, uncached/cached agreement is 0.510/0.297 ($p<10^{-8}$). Within the cached regime, reducing NFE four-fold leaves agreement at 0.297--0.300 (Friedman $p=0.544$; extreme-change CI [$-0.025,+0.030$]). Thus this second model supports the diagnosis: enabling prefix caching coincides with a large trajectory shift, while NFE does not control its magnitude in the tested range. LaViDa exposes no Fast-dLLM-style refresh interval, so this check supports the diagnosis, not the remedy.

\section{RQ4: Can Smarter Refresh Policies Beat the Fixed Interval?}
\label{sec:rq4}

The natural next step after RQ2 is an \emph{adaptive} refresh policy: monitor a cheap online signal and refresh only when needed, hoping to beat the fixed-interval frontier of Table~\ref{tab:refresh}. We tested five variant families and 12 configurations; the outcome is uniformly negative. Table~\ref{tab:negative} reports seven representative or strongest points plotted as grey $\times$'s in Figure~\ref{fig:teaser}b, each paired with the fixed-interval setting it must beat; throughout, ``matched compute'' means the fixed-interval reference has equal or lower measured wall-clock per generation on identical hardware.

\begin{sloppypar}
\textbf{Variants tested.} All variants run on the same 50 images at \texttt{threshold=0.5}, modifying only \emph{when} or \emph{how} the cache is refreshed. The complete grid is: (i)~\emph{token budget}, refreshing after $B \in \{4,8,16,32\}$ committed tokens; (ii)~\emph{confidence decay}, refreshing when mean commit confidence falls below $r \in \{0.85,0.95\}$ of its post-refresh level; (iii)~\emph{attention decay}, applying the same $r \in \{0.85,0.95\}$ rule to image-token attention mass at a fixed middle layer; (iv)~\emph{chunked rotation}, recomputing one of $C \in \{16,8,4\}$ image-token chunks per step; and (v)~one \emph{joint per-step recomputation} configuration, recomputing all image and generated-text positions while freezing only the 47-token generic system prefix. Table~\ref{tab:negative} reports the highest-agreement token-budget and decay points, all three chunk sizes, and the joint point; the unreported settings were also included in the statistical tests below.
\end{sloppypar}

\begin{table}[t]
  \caption{RQ4: seven reported adaptive/smoothed points (of 12 tested), each against a fixed-interval setting of equal or greater speed. No variant beats the frontier; exact-match counts are 0/50 except joint per-step (7/50, vs.\ 40/50 for interval 1).}
  \label{tab:negative}
  \footnotesize
  \setlength{\tabcolsep}{3.5pt}
  \begin{tabular}{llccl}
    \toprule
    Variant & Config & Time (s) & Jaccard & Fixed-int.\ ref. \\
    \midrule
    Token budget    & $B{=}4$      & 18.6 & 0.492 & int.\,4: 17.5s, 0.572 \\
    Conf.\ decay    & $r{=}0.85$   & 8.5  & 0.471 & int.\,16: 8.1s, 0.463 \\
    Attn.\ decay    & $r{=}0.95$   & 7.3  & 0.437 & int.\,16: 8.1s, 0.463 \\
    Chunked         & $C{=}16$     & 11.2 & 0.405 & int.\,8: 11.2s, 0.482 \\
    Chunked         & $C{=}8$      & 14.4 & 0.436 & int.\,8: 11.2s, 0.482 \\
    Chunked         & $C{=}4$      & 20.1 & 0.472 & int.\,4: 17.5s, 0.572 \\
    Joint per-step  & prefix frozen& 50.6 & 0.616 & int.\,1: 55.3s, 0.987 \\
    \bottomrule
  \end{tabular}
\end{table}

\textbf{Results.} The three signal-triggered policies all fail against the fixed interval at matched compute: paired differences are indistinguishable (all $p>0.4$, confidence decay) or significantly \emph{worse} (token budget $p=0.024$; attention decay $p=0.026$). The token-budget failure is mundane (near-constant ${\approx}1.1$ tokens/step makes token and step counts interchangeable), but the other two are informative: neither the model's confidence nor its attention to the image carries a usable early-warning signal for drift. The two smoothed-refresh architectures also fail. Chunked rotation is strictly dominated at every matched budget (losing $0.046$--$0.100$ Jaccard, paired $p \le 0.057$ at $C{=}16/8/4$), and joint per-step recomputation -- at compute comparable to \texttt{prefix\_refresh\_interval=1} (50.6s vs.\ 55.3s) -- reaches only Jaccard $0.616$ with $7/50$ exact matches, versus $0.987$ and $40/50$ (paired $p < 10^{-7}$).

\textbf{A feasibility probe for learned gating.} Could a \emph{trained} gating policy succeed where hand-designed triggers fail? Across 160 runs with \emph{randomized} refresh schedules (20 images $\times$ 8), no logged trajectory feature predicts final agreement post hoc (cross-validated $R^2\!\approx\!0.04$ from schedule statistics; $\approx\!0$ with all features). Most tellingly, re-running the \emph{same} image at the \emph{same} average refresh rate with a different realization of refresh \emph{times} shifts final Jaccard by $0.11$ on average (max $0.53$) -- nearly the full between-run standard deviation ($0.146$) -- leaving essentially nothing for a gating network to learn from.

\textbf{Interpretation.} Together these results support a simple reading: generation under confidence-aware parallel decoding is highly sensitive to the exact realization of the refresh schedule, and reproducing the reference trajectory is brittle in a near all-or-nothing way -- freezing 1.2\% of the sequence collapses exact reproduction from $40/50$ to $7/50$; adding an alternating image/text decomposition collapses agreement the rest of the way into the stale-cache band ($0.41$--$0.47$). Across every variant we designed, agreement tracked one quantity only: how close the per-step computation was to \emph{exactly} the reference computation -- never the cleverness of when, or how smoothly, an approximation was applied. This gives the refresh interval's dominance a structural explanation: among the parameters studied, it is the only one that buys agreement by making a fraction of steps exactly right rather than by introducing a new approximation whose deviation compounds. This conclusion covers the signal-triggered and smoothed-refresh families tested here; it cannot rule out fundamentally different designs (trained gating over richer features, retrieval-style re-grounding). Within the design space we explored, the fixed interval was never beaten, and we propose it as the frontier for future work to beat.

\section{Qualitative Analysis and Manual Verification}
\label{sec:audit}

\begin{sloppypar}
Word-level Jaccard is coarse, so we manually classify 50 low-similarity, image-deduplicated pairs as \emph{content substitution}, \emph{benign paraphrase}, or \emph{mixed}. This targeted-tail, single-author audit finds 25/50 (50\%) substitutions, 8/50 (16\%) paraphrases, and 17/50 (34\%) mixed; the substitution rate is stable between a 15-case pilot (47\%) and 35-case extension. Examples include a Roman street becoming unrelated ruins and an invented Hebrew translation, while ``Fenders Dinner'' versus ``Fender's Dinner'' is benign. Thus Jaccard conflates errors with rewording, but is not merely measuring lexical diversity; the blinded study below separately evaluates factuality against images.
\end{sloppypar}

\textbf{Semantic corroboration.} As a second, non-lexical check, we compute BERTScore-F1 (RoBERTa-large, rescaled) on the paired 50-image subset across all three sweeps (900 generations). It reproduces every qualitative finding: threshold-insensitivity (flat at $0.568$--$0.576$; Friedman $p=0.76$), monotonic Fast-dLLM recovery ($0.576 \to 0.991$; Page's $L$, $p<10^{-10}$), and dLLM-Cache's small, insufficient recovery ($0.407 \to 0.48$; Friedman $p=0.006$).

\textbf{Blinded image-grounded evaluation.} The audit above classifies \emph{differences} between paired outputs; it does not measure which side is right about the image. Two independent computer-science undergraduate annotators, neither involved in method development or experiments, therefore evaluated all 50 image pairs (baseline vs.\ default acceleration) with system identities hidden, A/B order randomized per item, and judgments made against the image. They recorded a five-way verdict and the number of image-inconsistent claims in each description. The proportion containing at least one such claim is $32\%/32\%$ (annotator 1) and $54\%/52\%$ (annotator 2) for baseline/acceleration. For each image, we average the two annotators' accelerated-minus-baseline error-count differences and test these 50 paired image-level averages. The mean is $0.00$ (95\% bootstrap CI $[-0.17,+0.17]$; Wilcoxon $p=0.97$); strictly-better vs.\ strictly-worse verdicts are $7{:}7$ and $8{:}7$. Five-way agreement is low (Cohen's $\kappa=0.18$; $0.21$--$0.26$ after binarization). Thus this study detects no factual-error difference at the default setting, but its interval and annotator disagreement do not establish factual equivalence or rule out moderate effects. Our main result remains about consistency with unaccelerated behavior.

\textbf{Ground-truth spot-check.} Five low-similarity cases show why drift is not error. The baseline is better on an illegible Hebrew plaque (the accelerated output invents a translation); acceleration is better on a ruins painting and church figures; and both outputs misidentify cities in two cases. ``Drift from baseline'' therefore must not be read as ``drift from truth'': our diagnosis concerns how content changes, not which side is more accurate.

\section{Discussion and Limitations}
\label{sec:discussion}

\begin{table}[t]
  \caption{Experiment ledger. RQ1 uses 300 paired baselines; targeted studies reuse its fixed 50-image subset.}
  \label{tab:ledger}
  \footnotesize
  \setlength{\tabcolsep}{3.5pt}
  \begin{tabular}{lrr}
    \toprule
    Experiment & Imgs $\times$ configs & Gens \\
    \midrule
    Unaccelerated baseline (RQ1 reference) & $300 \times 1$ & 300 \\
    RQ1 threshold sweep (Sec.~\ref{sec:rq1}) & $300 \times 5$ & 1{,}500 \\
    RQ1 threshold instrumentation (validation) & $50 \times 5$ & 250 \\
    Prompt/sample generalization & $50 \times 2 \times 5$ & 500 \\
    RQ2 refresh-interval sweep & $50 \times 6$ & 300 \\
    RQ2 state-refresh ablation / image swap & $50 \times 2 + 10 \times 4$ & 140 \\
    Threshold$\times$refresh grid (Sec.~\ref{sec:interaction}) & $50 \times 4$ & 200 \\
    RQ3 dLLM-Cache sweep (Sec.~\ref{sec:generalization}) & $50 \times 6$ & 300 \\
    RQ3 dLLM-Cache joint pilot & $15 \times 4$ & 60 \\
    LaViDa cross-model sweep (Sec.~\ref{sec:lavida}) & $50 \times 6$ & 300 \\
    RQ4 token budget & $50 \times 4$ & 200 \\
    RQ4 confidence / attention decay & $50 \times 4$ & 200 \\
    RQ4 chunked + joint & $50 \times 4$ & 200 \\
    RQ4 learned-gating probe & $20 \times 8$ & 160 \\
    Short-form (VQA) pilot & $20 \times 4$ & 80 \\
    Cross-hardware timing replication & $30 \times 8$ & 240 \\
    \midrule
    Total & & 4{,}930 \\
    \bottomrule
  \end{tabular}
\end{table}

\begin{sloppypar}
\textbf{Practical guidance.} For output stability, treat KV-cache refresh interval rather than confidence threshold as the primary lever (Table~\ref{tab:refresh}): the default gives $10$--$12\times$ speedup, while conservative intervals improve reproducibility and still retain $1.3\times$. This is not a safety control; it reproduces the baseline's errors as well as its behavior (Section~\ref{sec:audit}). More generally, sweeping an exposed hyperparameter against a paired unaccelerated reference costs only a few hundred generations and should precede deployment.
\end{sloppypar}

\textbf{Scope.} Table~\ref{tab:ledger} itemizes 4{,}930 generations (4{,}690 on RTX 6000 Ada; 240 A800 timing replications). RQ1 uses 300 images; costlier mechanism and generalization studies use paired 50-image subsets.

We note several limitations.

\begin{sloppypar}
\emph{First}, causal/remedy studies use LLaDA-V; LaViDa supports the diagnosis but lacks the refresh knob needed to test the remedy. Prompt generalization covers two additional single-turn formats on an independent 50-image sample, and a 20-image MME yes/no pilot points the same way, but larger-scale and multi-turn tests remain future work.
\end{sloppypar}

\emph{Second}, Jaccard is lexical; a 50-case audit and BERTScore corroborate each qualitative finding (Section~\ref{sec:audit}), but larger model-judged studies would strengthen it.

\begin{sloppypar}
\emph{Third}, dLLM-Cache transfers the diagnosis but requires tightening both caches to recover agreement, eliminating its speed advantage (Section~\ref{sec:generalization}).
\end{sloppypar}

\begin{sloppypar}
\emph{Fourth}, agreement is not ground truth. Our blinded image-grounded study covers only the default configuration and 50 images (detectable effect roughly $\pm0.17$ errors/description), with low five-way inter-annotator agreement ($\kappa=0.18$); broader factuality evaluation remains future work.
\end{sloppypar}

\begin{sloppypar}
\emph{Fifth}, our Web-facing motivation is not a Web-specific evaluation. MME is a general multimodal benchmark; we do not test rendered webpages, search or ranking outcomes, product/news pages, or end-to-end agent decisions. Those settings motivate the diagnostic but are not measured deployment claims.
\end{sloppypar}

\section{Conclusion}
Within the parallelism range studied, confidence threshold does not control acceleration-induced content drift despite being demonstrably operative. KV-cache refresh interval is the dominant observed consistency control for the tested Fast-dLLM implementation, providing a continuous speed--agreement trade-off; dLLM-Cache reproduces the initial diagnosis but not the efficient remedy, and LaViDa provides an auxiliary cross-model check. Our target is consistency with unaccelerated behavior, not factuality: the limited blinded study detects no factual-error difference but does not establish equivalence. The paired diagnostic may be useful for evaluating other accelerated generators.

\section*{Ethical Considerations}
This work analyzes publicly released, pretrained multimodal models and the public MME benchmark, and releases no new model weights. Human involvement consisted of a bounded evaluation by two computer-science undergraduates who were independent of method development and experiment execution. They evaluated model-generated descriptions against public benchmark images; the study analyzes the model outputs, not the annotators. We draw no conclusions about annotator behavior, use no demographic or personal attributes as study variables, and report judgments only in aggregate. Our drift reference is the model's own unaccelerated output, not ground truth, so high agreement must not be interpreted as factual reliability. Likewise, refresh interval is a consistency control, not a safety control: conservative refresh reproduces baseline behavior, including its errors. Deployment in medical, legal, accessibility, surveillance, or other consequential settings requires task-specific factuality evaluation and appropriate human oversight regardless of acceleration settings. Our findings are diagnostic; we are not aware of additional dual-use risk from characterizing this existing acceleration behavior.

\bibliographystyle{ACM-Reference-Format}
\bibliography{references}

\end{document}